\pdfoutput=1

\documentclass[11pt]{article}

\usepackage{emnlp2021}

\usepackage{times}
\usepackage{latexsym}

\usepackage[T1]{fontenc}

\usepackage[utf8]{inputenc}

\usepackage{microtype}

\usepackage{natbib}
\usepackage{graphicx}
\usepackage{siunitx}
\usepackage{booktabs}
\usepackage{longtable}
\usepackage{caption}
\usepackage[]{algorithm2e}
\usepackage[normalem]{ulem}
\usepackage{float}
\newtheorem{theorem}{Theorem}

\usepackage{subcaption}

\usepackage{enumitem}
\setitemize{noitemsep,topsep=0pt,parsep=0pt,partopsep=0pt}
\usepackage{amsfonts}
\usepackage{chngcntr}
\usepackage{multirow}

\usepackage{amsmath}
\usepackage{bbold}

\DeclareMathOperator*{\E}{\mathbb{E}}

\newif{\ifhidecomments}

\ifhidecomments
    \newcommand{\jdcomment}[1]{}	
    \newcommand{\nascomment}[1]{}
    \newcommand{\dallas}[1]{}
    \newcommand{\roy}[1]{}
    \newcommand{\suchin}[1]{}
    \newcommand{\roys}[1]{}
    \newcommand{\resolved}[1]{}
\else
    \newcommand{\jdcomment}[1]{\textcolor{teal}{[#1 ({\bf Jesse})]}} 
    \newcommand{\nascomment}[1]{\textcolor{red}{[#1 ({\bf Noah})]}} \newcommand{\dallas}[1]{\textcolor{orange}{[#1 ({\bf DBC})]}} 
    \newcommand{\suchin}[1]{\textcolor{cyan}{[#1 ({\bf SG})]}} 
    \newcommand{\roy}[1]{\textcolor{blue}{[#1 ({\bf Roy})]}} 
     
    \newcommand{\roys}[1]{\textcolor{blue}{\sout{#1}}} 
    \newcommand{\resolved}[1]{} 
\fi

\title{Expected Validation Performance \\ and Estimation of a Random Variable's Maximum}

\newcommand{\aspace}{\hspace{1em}}
\newcommand{\uw}{$^{\heartsuit}$}
\newcommand{\aiTwo}{$^{\clubsuit}$}
\newcommand{\stanford}{$^{\spadesuit}$}
\newcommand{\hebrew}{$^{\diamondsuit}$}

\author{Jesse Dodge\aiTwo \aspace
Suchin Gururangan\uw \aspace
Dallas Card\stanford \aspace \\
\textbf{Roy Schwartz}\hebrew \aspace
\textbf{Noah A. Smith}\uw\aiTwo \aspace\\
\stanford Stanford University \\
\hebrew Hebrew University of Jerusalem \\
\uw Paul G.\ Allen School of Computer Science \& Engineering, University of Washington \\
\aiTwo Allen Institute for Artificial Intelligence\\
\texttt{jessed@allenai.org}
}

\begin{document}
\maketitle
\begin{abstract}
Research in
NLP is often supported by experimental results, and improved reporting of such results can lead to better understanding and more reproducible science. In this paper we analyze three statistical estimators for expected validation performance, a tool used for reporting performance (e.g., accuracy) as a function of computational budget (e.g., number of hyperparameter tuning experiments). Where previous work analyzing such estimators focused on the bias, we also examine the variance and mean squared error (MSE). In both synthetic and realistic scenarios, we evaluate three estimators and find the unbiased estimator has the highest variance, and the estimator with the smallest variance has the largest bias; the estimator with the smallest MSE strikes a balance between bias and variance, displaying a classic bias-variance tradeoff. We use expected validation performance to compare between different models, and analyze how frequently each estimator leads to drawing incorrect conclusions about which of two models performs best. We find that the two biased estimators lead to the fewest incorrect conclusions, which hints at the importance of minimizing variance and MSE.
\end{abstract}

\section{Introduction}

Drawing robust conclusions when comparing different methods in natural language processing is central to scientific progress.
If two research groups set up the same set of experiments, they should expect to get similar results.
One area that has high impact, but is often underreported, is hyperparameter tuning \citep{score_distributions, damour2020underspecification, dodge-etal-2019-show, melis2018}. 
Hyperparameter search is key to getting strong results; for example, RoBERTa \citep{liu2019roberta} found stronger results than BERT \citep{bert} partly due to an increased budget for hyperparameter tuning.
Often researchers only report the performance of the single best-found model during a hyperparameter search \cite{Ethayarajh2020Utility, forde2019scientific, sculley2018}.
What if a future researcher has a smaller computational budget for training models? What performance should they expect to find?
One way of reporting such results is expected validation performance (EVP).

What is EVP? Assume a budget to train $B$ models (e.g., $B$ rounds of hyperparameter search), with resulting evaluation scores (e.g., accuracy) on the validation set $X_1\ldots X_B$. Standard practice would report the maximum result, $X_{max}$, but this effectively hides the experiments which were required to achieve that maximum performance. Using all $B$ results, EVP estimates what the maximum would have been if we had had a \textit{smaller} budget $n$ (where $1\leq n < B$).
This is estimating what the maximum of $n$ trials \emph{would be}, in expectation; this is thus a statistical estimation problem. The formulation was introduced by \citet{dodge-etal-2019-show}, who proposed a first estimator (defined as $V_n^B$ in Equation~\ref{eq:v_def}). This estimator was later shown to be biased by \citet{tang-etal-2020-showing}, who introduced an unbiased estimator (defined as $U_n^B$ in Equation~\ref{eq:u_def}) for the same expected maximum.

In Section~\ref{sec:estimation} we use tools from combinatorics to derive both previously-introduced estimators, relate them to each other, and show that they make two opposing assumptions; we show that changing only one of these assumptions instead of both leads to a third estimator, $W_n^B$, and prove that this estimator is even more biased than $V_n^B$. 

Unbiased estimators are generally preferred, all else equal, but only analyzing the bias provides an incomplete picture of the quality of an estimator. In Section~\ref{sec:sim} we also measure the variance and mean squared error of these three estimators in synthetic experiments. 
We find that while $U_n^B$ is unbiased (as expected) it has the highest variance, and that $W_n^B$ is the most biased but has the lowest variance; $V_n^B$ strikes a balance between the bias and variance, leading to the lowest mean squared error (so, the average squared distance to the true value being estimated is smallest).\footnote{There is a long tradition of preferring biased estimators over unbiased ones \citep{all_of_stats}, such as when estimating the population variance using the sample variance, or the James–Stein estimator \cite{steins_estimator}.}


Finally, in Section~\ref{sec:incorrect_conclusions} we explore how these estimators impact a common use case of EVP: comparing the results of hyperparameter searches for two models. Specifically, we examine how frequently the estimators lead to incorrectly concluding that the worse model outperforms the better one (for a given budget), and find that the high-variance (but unbiased) $U_n^B$ more frequently leads to such incorrect conclusions than the other lower variance (but biased) estimators.

\section{Estimation of the Expected Maximum}\label{sec:notation}

\def\Var{{\textrm{Var}}\,}
\def\E{{\mathbb{E}}\,}
\def\Bias{{\textrm{Bias}}\,}
\def\MSE{{\textrm{MSE}}\,}
\def\maxrv{{Y_n}\,}
\def\expmax{{\theta}}
\newcommand{\textmultiset}[2]{\bigl(\!{\binom{#1}{#2}}\!\bigr)}
\newcommand{\displaymultiset}[2]{\left(\!{\binom{#1}{#2}}\!\right)}
\newcommand\multiset[2]{\mathchoice{\displaymultiset{#1}{#2}}
                                {\textmultiset{#1}{#2}}
                                {\textmultiset{#1}{#2}}
                                {\textmultiset{#1}{#2}}}
\paragraph{Notation}
We begin by defining some notation. Consider $n$ i.i.d.~random variables, $X_1, \ldots, X_n\sim F$, for some unknown $F$.\footnote{For clarity, we dispense with notation mapping into the use case of interest, as well as the computational details; see \citet{dodge-etal-2019-show} for a full discussion.}
\begin{itemize}
    \item $\maxrv=\max\{X_1,\ldots,X_n\}$, a random variable representing the maximum of $n$ i.i.d. random variables.
    \item $\expmax_n=\E[\maxrv]$, the true expected value of $\maxrv$.
    \item $\hat{\expmax}_n$, an estimator of $\expmax_n$ (the expected value).
    \item $\Bias(\hat{\expmax})=\E[\hat{\expmax}]-\expmax$, the bias of $\hat{\expmax}$. \item $\Var(\hat{\expmax})$, the estimator's variance due to sampling.
    \item $\MSE(\hat{\expmax})=\Bias(\hat{\expmax})^2+\Var(\hat{\expmax})$, the mean squared error of the estimator. MSE is the average squared difference between the estimator and true value, or the expected value of the squared error loss between the estimator and the true statistic.
\end{itemize}
\paragraph{Estimation of the Expected Maximum}\label{sec:estimation}
We consider the estimation of $\expmax$, the expected maximum. With a finite sample of $B$ draws from $F$, we can estimate this quantity for $1\leq n\leq B$. We begin with the definition of an expectation over a discrete set: $\E[\maxrv]=\sum_{i=1}^{B} X_i P(\maxrv=X_i)$.
This can be rewritten using order statistics. Let $X_{(i)}$ denote the $i$th largest sample (distinct from $X_i$). Then, 
\begin{align}\label{eq:exp_val}
    &\E[\maxrv]=\textstyle\sum_{i=1}^{B} X_{(i)} P(\maxrv=X_{(i)})\\
    &=\textstyle\sum_{i=1}^{B} X_{(i)} \left( P(\maxrv\leq X_{(i)}) - P(\maxrv< X_{(i)})\right)\nonumber\\
    &=\textstyle\sum_{i=1}^{B} X_{(i)} \left(P(\maxrv\leq X_{(i)}) - P(\maxrv\leq X_{(i-1)})\right)\nonumber
\end{align}
This estimation depends on $P(\maxrv\leq X_{(k)})$, the probability that a sample of size $n$ has a maximum that is less than or equal to the $k$th order statistic. We can estimate this probability by counting: from our $B$ points how many sets of size $n$ are there which only include order statistics up to $k$, out of the total number of sets of size $n$? We turn to combinatorics, which provides tools for counting such sets. Two key assumptions must be made: whether the sets will contain repetition or not and whether the items in the sets will be ordered or unordered. These assumptions will lead to different estimators.

Ordered subsets that allow repetition are known as \textbf{strings}, and there are $B^n$ strings of size $n$ from $B$ points. With these assumptions, we now have a closed form for $P(\maxrv\leq X_{(k)})$, and plugging this into Equation~\ref{eq:exp_val} we define our first estimator:
\begin{equation}\label{eq:v_def}
\begin{split}
    V_n^B&= \sum_{i=1}^{B} X_{(i)} \left(\frac{i^n}{B^n}-\frac{(i-1)^n}{B^n}\right).
\end{split}
\end{equation}
This is exactly the estimator introduced in \citet{dodge-etal-2019-show}, derived using the plug-in estimator for the CDF (the empirical CDF).

Making the opposite two assumptions, unordered subsets without repetition are \textbf{combinations}, for which there are $\binom{B}{n}$ subsets of size $n$ from $B$ points. The corresponding estimator is 
\begin{equation}\label{eq:u_def}
\begin{split}
    U_n^B&= \sum_{i=1}^{B} X_{(i)} \left(\frac{\binom{i}{n}}{\binom{B}{n}}-\frac{\binom{i-1}{n}}{\binom{B}{n}}\right).
\end{split}
\end{equation}
This is the estimator of \citet{tang-etal-2020-showing}, which they derived as an unbiased estimator.

What about changing only one of these assumptions?
Ordered subsets without repetition are \textbf{permutations}, for which there are $_{B} P_{n}$ subsets of size $n$ from $B$ points. Though these assumptions are different, the corresponding estimator is equivalent to $U_n^B$, since: 
\begin{equation}
    \frac{_{k} P_{n}}{_{B} P_{n}} =
    \frac{\frac{k!}{(k-n)!}}{\frac{B!}{(B-n)!}} =
    \frac{\frac{k!}{n!\,(k-n)!}}{\frac{B!}{n!\,(B-n)!}} =
    \frac{\binom{k}{n}}{\binom{B}{n}}
\end{equation}

Finally, unordered subsets with repetition are \textbf{multisets}, the number of which is denoted $\multiset{B}{n}= \binom{B+n-1}{n}$. We introduce the corresponding estimator:
\begin{equation}\label{eq:w_def}
\begin{split}
    W_n^B&= \sum_{i=1}^{B} X_{(i)} 
    \left(\frac{\multiset{i}{n}}{\multiset{B}{n}}-\frac{\multiset{i-1}{n}}{\multiset{B}{n}}\right).
\end{split}
\end{equation}

\paragraph{Comparing estimators}
To compare these estimators we turn to the standard statistical tools of bias, variance, and mean squared error. $U_n^B$ was shown to be unbiased, and $\Bias(V_n^B)\leq 0$ \citep{tang-etal-2020-showing}. We show that $\Bias(W_n^B)\leq \Bias(V_n^B)$, that is $W_n^B$ has a larger negative bias than $V_n^B$.
\begin{theorem}
Assume $X_1, \ldots, X_B\sim F$ are i.i.d.~from unknown distribution $F$. Let $1\leq k<B$, and $1 \leq n \leq B$. Then, $\Bias(W_n^B)\leq \Bias(V_n^B)$.
\end{theorem}
Consider $V_n^B$ as defined in Equation~\ref{eq:v_def}. The sum of the coefficients of the $X_{(i)}$ up to $k$ is  $\frac{k^n}{B^n}$. It is sufficient to show that, for a given $k$, this term is less than the sum of the coefficients for $W_n^B$, which is $\left.{\multiset{k}{n}}\right/{\multiset{B}{n}}$; this implies that $V_n^B$ places less probability mass on the smaller order statistics than $W_n^B$.
\begin{align}\label{eq:first_ineq}
    \frac{k^n}{B^n}< \frac{\multiset{k}{n}}{\multiset{B}{n}} & \iff \frac{k^n}{B^n} < \frac{\binom{k+n-1}{n}}{\binom{B+n-1}{n}} \\
    & \iff \frac{\binom{B+n-1}{n}}{B^n} < \frac{\binom{k+n-1}{n}}{k^n}. \label{eq:first_ineq_b}
\end{align}
The left side of Eq.~\ref{eq:first_ineq_b} can be rewritten as:
\begin{align}
    \frac{\binom{B+n-1}{n}}{B^n} &= \frac{\frac{(B+n-1)!}{n!\, B!}}{B^n} = \frac{1}{n!}\frac{\prod_{j=0}^{n-1}\left(B+n-1-j\right)}{B^n} \nonumber \\
    &=\textstyle \left(\frac{1}{n!}\right)\prod_{j=0}^{n-1}\left(1+\frac{(n-1)-j}{B}\right).
\end{align}
Rewriting the right side of Eq.~\ref{eq:first_ineq_b} in a similar manner, we have
\begin{align*}
\textstyle    \prod_{j=0}^{n-1}\left(1+\frac{(n-1)-j}{B}\right) < \prod_{j=0}^{n-1}\left(1+\frac{(n-1)-j}{k}\right)
\end{align*}
since $B>k$. This completes our proof.

\section{Simulation Experiment} \label{sec:sim}


In the previous section we proved that $W_n^B$ is at least as biased as $V_n^B$, but such a bound tells us little about how these estimators behave in practice. In this section we provide a simulation experiment which allows us to measure the bias and variance of each estimator directly.
We assume a distribution for $X_i$, which allows us to draw many samples of size $B$ so we can evaluate how these estimators behave.
Recall that the motivating application of our estimators is when $\{X_i\}_{i=1}^{n}$ represent the evaluations from different trials of hyperparameter optimization, so designing a reasonable distribution for $X_i$ allows us to evaluate the estimators with tens of thousands of simulated trials without having to train that many models.

\subsection{Synthetic Experiments Setup}\label{subsec:synthetic}
To begin, we sample 100,000 random values from a Normal$(0.6, 0.07)$ distribution (truncated to $[0, 1]$). We then sample 10,000 values from this set, resulting in 9536 unique values, with a true maximum of 0.854. Call this bag of values $\mathcal{V}$.
We then set $B=30$, and estimate the true EVP as a function of $n$ for $n=1,...,30$, by drawing 50,000 samples of size $n$ from $\mathcal{V}$, for each value of $n$, and reporting the average maximum for each $n$ (``True EVP'' in Figure \ref{fig:synthetic_bias_and_var}, top).
To estimate the mean and variance of a given estimator we sample 10,000 $B$ values from $\mathcal{V}$ and compute the value of the estimator for each, then calculate the mean and variance across those 10,000 samples.

\subsection{Bias, Variance, MSE}

\begin{figure}[t]
    \centering
    \includegraphics[scale=0.64]{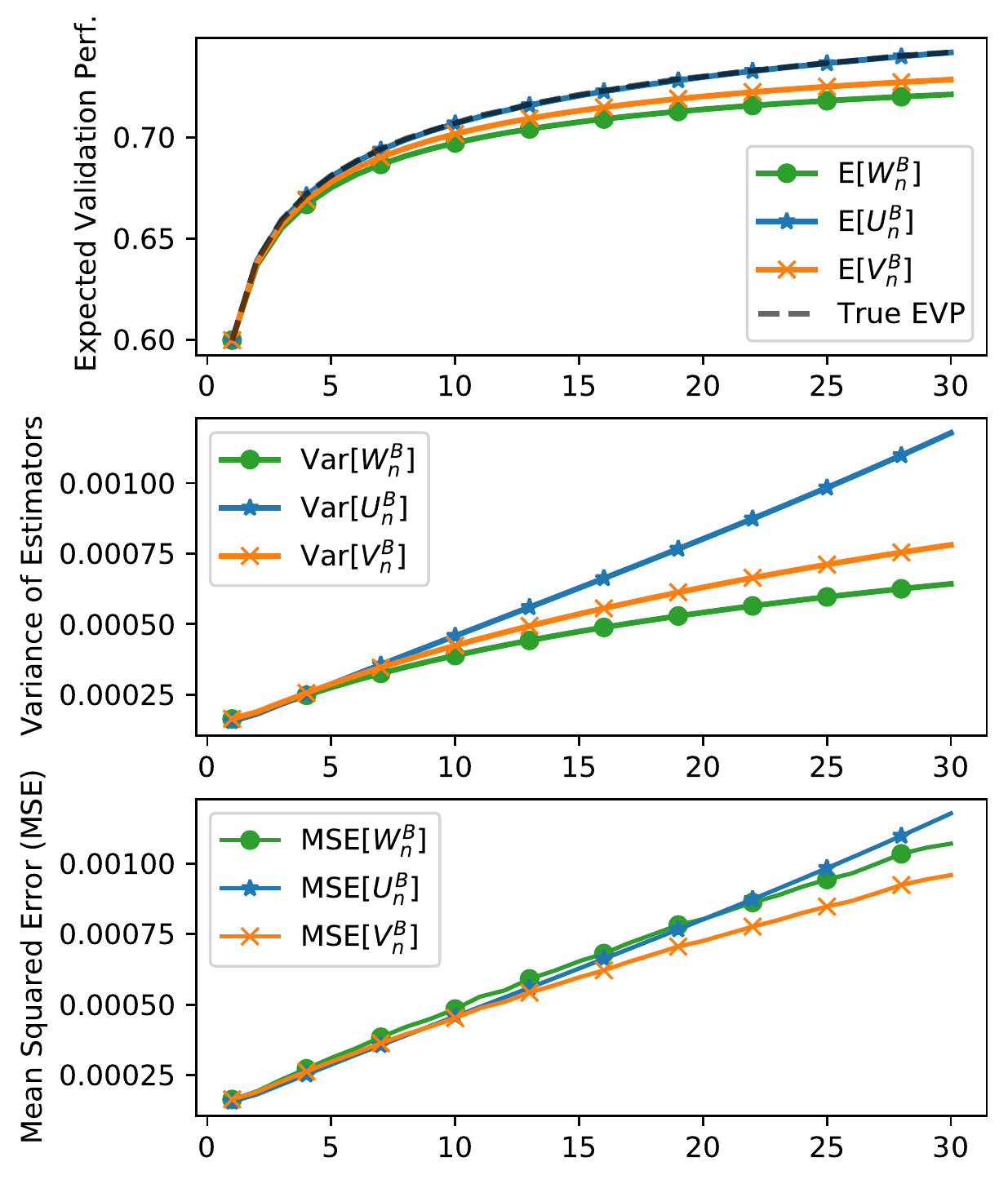}
    \caption{Expected value (top) variance (middle) and mean squared error (MSE; bottom) of the three estimators, based on a 10,000 simulations for a large random bag of possible validation scores. As expected, $U_n^B$ is unbiased while $V_n^B$ and $W_n^B$ have negative bias. However, $W_n^B$ has the lowest variance, and $V_n^B$ 
    balances bias and variance, leading to lowest MSE.}
    \label{fig:synthetic_bias_and_var}
\end{figure}

Figure \ref{fig:synthetic_bias_and_var} shows the estimated mean (top), variance (middle), and MSE (bottom) of each estimator. As can be seen in the top figure, $\Bias(W_n^B)\leq \Bias(V_n^B)\leq \Bias(U_n^B)=0$ with a a difference that grows with $n$, confirming the proved bounds for these estimators. In the middle figure we measure the variance of these estimators, and we see that $\Var(W_n^B)\leq \Var(V_n^B)\leq \Var(U_n^B)$, with the difference in variance again growing with $n$.

In the bottom of Figure~\ref{fig:synthetic_bias_and_var} we plot the mean squared error (MSE); as a reminder, 
$\MSE(\hat{\expmax})=\Bias(\hat{\expmax})^2+\Var(\hat{\expmax})$, so lower is better. Although $U_n^B$ is unbiased, and $W_n^B$ has the lowest variance, $V_n^B$ strikes the balance between bias and variance that leads to the lowest MSE.

Thus we see that a higher variance estimator may, on average, be farther from the true value than a biased but lower variance estimator. Again tying this back to our motivating application of hyperparameter tuning, in this scenario $V_n^B$ is more likely to underestimate than overestimate performance for a given budget, but overall will have lower variance between researchers running sets of experiments, and will on average have closer predictions to the true value than the other two estimators.

\section{Incorrect Conclusions}\label{sec:incorrect_conclusions}

While analyzing how close each estimator is to the true expected maximum for one model is important, in practice these curves are often used to compare two or more different models. For example, NLP practitioners may run hyperparameter searches for two different models, compute the expected validation curves for each, and select the model which presents a higher estimated maximum performance \cite{Zhang2021revisiting, gehman2020realtoxicityprompts}.
In this section we examine the three estimators in such a scenario, asking how frequently each estimator leads to drawing \emph{incorrect} conclusions about which model performs best for a specific budget.


\subsection{Experimental Setup}
We proceed by performing a sensitivity analysis: we run 100 trials of random hyperparameter search (far more than is typically necessary to establish that one model outperforms another in current practice) for a CNN \cite{kim2014convolutional} and a linear bag-of-embedding (LBoE) \cite{Yogatama2015BayesianOO}. These models are trained on the Stanford sentiment treebank 5-way text classification task \cite{Socher2013RecursiveDM}. We include details about the dataset (and a link to download it) in Appendix~\ref{app:data}.

For all three estimators, the CNN has higher expected performance than the LBoE, for all $n \le B$.\footnote{See Appendix~\ref{app:real_exp} for details.
Figure~\ref{fig:evp_real_data} shows expected validation curves for $B=100$ for all three estimators; with $B$ this large, the three estimators are very similar.}
We then simulate a more practical scenario where a practitioner runs $B \in \{15, \ldots, 30\}$ rounds of hyperparameter search for the two models and compares their estimated maximum at $n=B$ (so, the estimated maximum of $B$ points) to conclude which is best (that is, which estimator has lower error).

We are interested in the rate at which each estimator would draw an incorrect conclusion about which model performs best. To evaluate this question we do the following: for each value of $B$ we sample 50,000 times from the 100 real experiments and compute the fraction for which the value of each estimator for the CNN is less than for LBoE. For example, to estimate the proportion with which $U_B^B$ draws an incorrect conclusion with $B=15$ we draw 50,000 samples of size 15 from the 100 real experimental results for each of the CNN and LBoE, then compute the fraction of those samples for which $U_B^B$ for the CNN is less than $U_B^B$ for LBoE.
A stable estimator will make the same prediction with small and large $B$.

\subsection{Results}

\begin{figure}[t]
    \centering
    \includegraphics[width=0.5\textwidth]{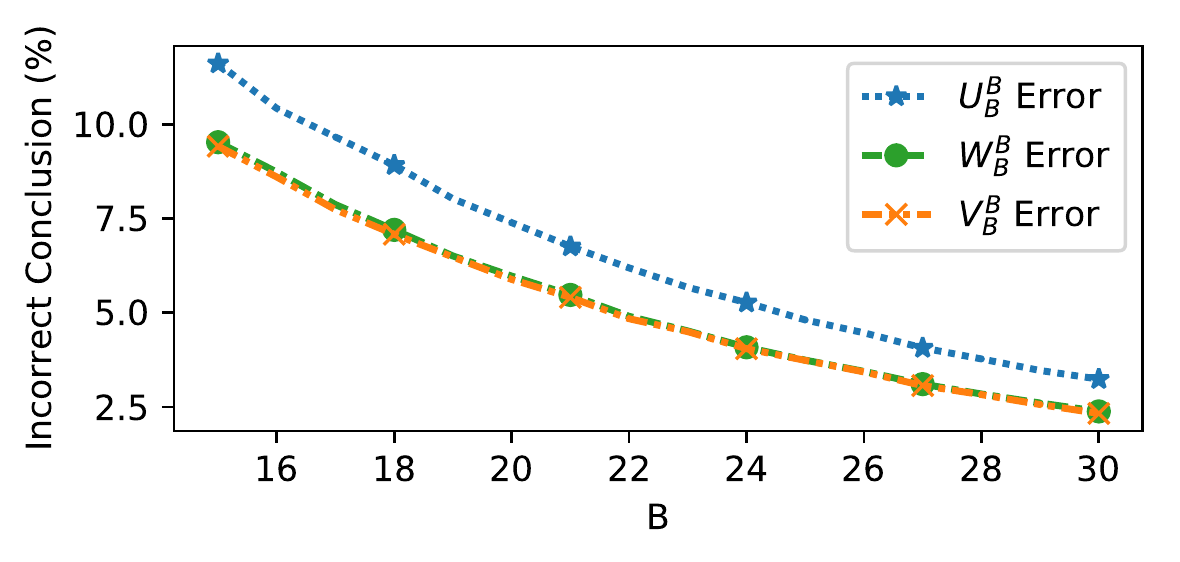}
    \caption{For a budget of $B$ trials, what fraction of the time does each estimator \emph{incorrectly} predict that the expected maximum of those $B$ trials (so, $n=B$) is higher for the LBoE than for the CNN? Lower is better.
    The proportion of errors made by the unbiased estimator $U_n^B$ when $n=B$ is higher than for either of the biased estimators, $V_n^B$ and $W_n^B$. Confidence intervals around this proportion are not shown, as they are small.
    \label{fig:error}}
\end{figure}

In Figure~\ref{fig:error} we see the results of this experiment: $U_B^B$ more frequently would lead a practitioner to incorrectly conclude that the LBoE outperforms the CNN for budgets $B\in \{15, \ldots, 30\}$ than $V_B^B$ or $W_B^B$. This scenario models what we expect a practitioner would care about: the frequency with which one draws conclusions that would be consistent with conclusions drawn with a larger budget. Here the high variance of $U_B^B$ likely plays a role the stability of its predictions; while it may be unbiased, the lower variance estimators are more reliable.

\section{Conclusion}


Drawing reproducible conclusions from our experimental results is of paramount importance to NLP researchers, practitioners, and users of language technologies. 
Expected validation performance curves are tools for comparing the results of hyperparameter searches; we showed how two previously-introduced estimators are connected through combinatorial assumptions, and introduced a third estimator by varying such assumptions. 
In synthetic experiments, we analyzed the bias, variance, and mean squared error, and found a classic example of a bias-variance tradeoff; the unbiased estimator $U_n^B$ had the largest variance, and the most biased estimator $W_n^B$ had the lowest variance, while $V_n^B$ struck a balance leading to the lowest mean squared error. 
Finally, in realistic experiments we found that the unbiased estimator led to incorrectly identifying the better of two models at a higher rate than the lower variance estimator.
Overall, $V_n^B$ had the lowest MSE and the lowest rate of drawing incorrect conclusions, so $V_n^B$ is our recommendation for estimating the expected maximum.

\section*{Acknowledgements}
Dallas Card was supported in part by the Stanford Data Science Institute.

\bibliography{anthology,acl2021}
\bibliographystyle{acl_natbib}

\newpage
\appendix

\section{Expected Validation Curves for two models, all three estimators}

\begin{figure*}
    \centering
    \includegraphics[scale=.65]{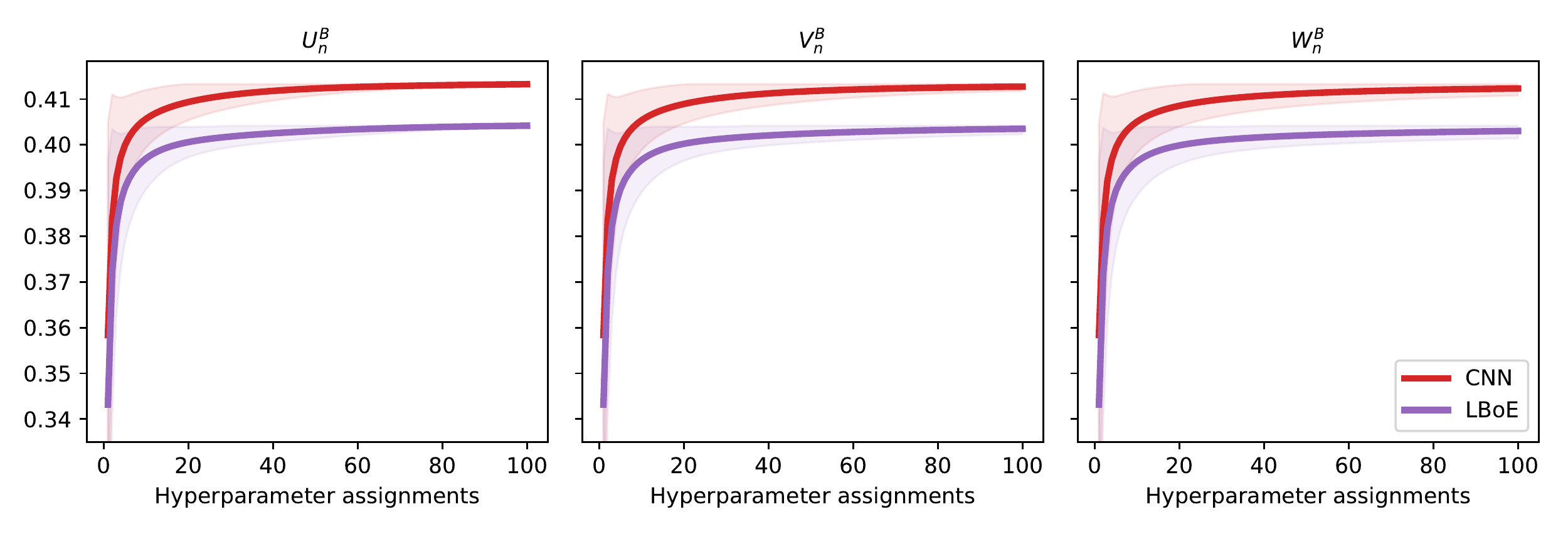}
    \caption{$U_n^B$ (left), $V_n^B$ (middle) and $W_n^B$ (right) curves of the same data, a CNN and a Linear Bag of Embeddings (LBoE), evaluated on SST-5, with $B$=100. With such a large $B$ the three estimators are very similar. For all three estimators, the CNN has higher expected performance than the LBoE for all $n$.}
    \label{fig:evp_real_data}
\end{figure*}

We include expected validation curves of the same data using all three estimators in Figure~\ref{fig:evp_real_data}. They look roughly the same.

\section{Training Data}\label{app:data}
The CNN and LBoE in Section~\ref{sec:incorrect_conclusions} are trained on the Stanford sentiment treebank 5-way text classification task \cite{Socher2013RecursiveDM}.
There are 8544 train examples, 2210 test examples, and 1101 validation examples. It can be downloaded here: \url{http://nlp.stanford.edu/sentiment}. We present label distributions in Table~\ref{app_table:dataset_statistics}.

\begin{table}
    \centering
    \begin{tabular}{c|ccc}
    Label & Train & Valid & Test\\ 
    \hline
    0 & 1092 & 139 & 279\\
    1 & 2218 & 289 & 633\\
    2 & 1624 & 229 & 389\\
    3 & 2322 & 279 & 510 \\
    4 & 1288 & 165 & 399\\
    \end{tabular}
    \caption{Label distributions for SST-5. 0 is ``very negative'', 2 is ``neutral'', and 4 is ``very positive''.}
    \label{app_table:dataset_statistics}
\end{table}

\section{Hyperparameter Ranges}\label{app:real_exp}
The hyperparameter bounds for the CNN and LBoE in Section~\ref{sec:incorrect_conclusions}, which were trained on SST-5 as described in Appendix~\ref{app:data}.

\begin{table*}[h]
    \centering
    \small
    \begin{tabular}{cc}
       \toprule
       \textbf{Computing infrastructure} & GeForce GTX 1080 GPU\\ 
       \midrule
       \textbf{Number of search trials} & 100 \\
       \midrule
       \textbf{Search strategy} & uniform sampling \\
       \midrule
       \textbf{Best validation accuracy} & 41.3\\
       \midrule
       \textbf{Training duration} & 77 sec\\
       \bottomrule
    \end{tabular}

    \vspace{3mm}\begin{tabular}{cccccccccccccc}
    \toprule
    
    \textbf{HP} & number of epochs & patience & batch size & embedding & encoder & max filter size  \\
    \midrule
    \textbf{Search space}  & 50 & 10 & 64 & GloVe (50 dim) & Convnet &  \emph{uniform-integer}[1, 9]  \\
    \midrule
    \textbf{Best assignment} & 50 & 10 & 64 & GloVe (50 dim) & Convnet & 9 \\
    
    \end{tabular}

    \vspace{3mm}\begin{tabular}{cccccccccccccc}
    \toprule
    
    \textbf{HP} &  number of filters &  dropout & LR scheduler & patience & reduction factor  \\
    \midrule
    \textbf{Search space}  &  \emph{uniform-integer}[64, 512] & \emph{uniform-float}[0, 0.5] & reduce on plateau & 2 epochs & 0.5  \\
    \midrule
    \textbf{Best assignment}   & 390 &  0.2 & reduce on plateau & 2 epochs &  0.5
    
    \end{tabular}

        \vspace{3mm}\begin{tabular}{cccccccccccccc}
    \toprule
    
    \textbf{HP} & optimizer & LR \\
    \midrule
    \textbf{Search space}  &  Adam & \emph{loguniform-float}[1e-6, 1e-1] \\
    \midrule
    \textbf{Best assignment}   &  Adam & 0.0004
    
    \end{tabular}

    \caption{SST (fine-grained) CNN classifier search space and best assignments.}
    \label{tab:my_label}
    
\end{table*}

\begin{table*}[h]
    \centering
    \small
    \begin{tabular}{cc}
       \toprule
       \textbf{Computing infrastructure} & GeForce GTX 1080 GPU\\ 
       \midrule
       \textbf{Number of search trials} & 100 \\
       \midrule
       \textbf{Search strategy} & uniform sampling \\
       \midrule
       \textbf{Best validation accuracy} & 42.7\\
       \midrule
       \textbf{Training duration} & 41 sec\\
       \bottomrule
    \end{tabular}

    \vspace{3mm}\begin{tabular}{ccccccccccc}
    \toprule
    
    \textbf{Hyperparameter} & number of epochs & patience & batch size & embedding & dropout  \\
    \midrule
    \textbf{Search space}  & 50 & 10 & 64 & GloVe (50 dim) & \emph{uniform-float}[0, 0.5]  \\
    \midrule
    \textbf{Best assignment} & 50 & 10 & 64 & GloVe (50 dim) & 0.4 \\
    \midrule
    \end{tabular}

        \vspace{3mm}\begin{tabular}{ccccccccccc}
    \toprule
    
    \textbf{Hyperparameter} &  LR scheduler & patience & reduction factor &  optimizer & LR \\
    \midrule
    \textbf{Search space}  & reduce on plateau & 2 epochs & 0.5 & Adam & \emph{loguniform-float}[1e-6, 1e-1] \\
    \midrule
    \textbf{Best assignment} & reduce on plateau & 2 epochs &  0.5 & Adam & 0.044\\
    \midrule
    
    \end{tabular}

    \caption{SST (fine-grained) BOE classifier search space and best assignments.}
    \label{tab:my_label}
    
\end{table*}

\end{document}